\documentclass[conference]{IEEEtran}
\IEEEoverridecommandlockouts
\usepackage{cite}
\usepackage{amsmath,amssymb,amsfonts}
\usepackage{algorithmic}
\usepackage{graphicx}
\usepackage{float}
\usepackage{hyperref}
\usepackage{fancyhdr}
\usepackage{lipsum} 

\fancypagestyle{plain}{
  \fancyhf{}
  \fancyfoot[C]{\thepage}
  }
\usepackage{textcomp}
\usepackage{xcolor}
\def\BibTeX{{\rm B\kern-.05em{\sc i\kern-.025em b}\kern-.08em
    T\kern-.1667em\lower.7ex\hbox{E}\kern-.125emX}}
\begin{document}

\title{Novel View Synthesis with Gaussian Splatting: Impact on Photogrammetry Model Accuracy and Resolution}

\author{
    \IEEEauthorblockN{Pranav Chougule}
    \IEEEauthorblockA{
        \textit{MS in Robotics \& Autonomous Systems} \\
        \textit{School for Engineering of Matter, Transport \& Energy} \\
        \textit{Arizona State University}\\
        Tempe, USA \\
        pchougu1@asu.edu
    }
}
\maketitle
\begin{abstract}
In this paper, I present a comprehensive study comparing Photogrammetry and Gaussian Splatting techniques for 3D model reconstruction and view synthesis. I created a dataset of images from a real-world scene and constructed 3D models using both methods. To evaluate the performance, I compared the models using structural similarity index (SSIM), peak signal-to-noise ratio (PSNR), learned perceptual image patch similarity (LPIPS), and lp/mm resolution based on the USAF resolution chart. A significant contribution of this work is the development of a modified Gaussian Splatting repository, which I forked and enhanced to enable rendering images from novel camera poses generated in the Blender environment. This innovation allows for the synthesis of high-quality novel views, showcasing the flexibility and potential of Gaussian Splatting. My investigation extends to an augmented dataset that includes both original ground images and novel views synthesized via Gaussian Splatting. This augmented dataset was employed to generate a new photogrammetry model, which was then compared against the original photogrammetry model created using only the original images. The results demonstrate the efficacy of using Gaussian splatting to generate novel high-quality views and its potential to improve photogrammetry-based 3D reconstructions. The comparative analysis highlights the strengths and limitations of both approaches, providing valuable information for applications in extended reality (XR), photogrammetry, and autonomous vehicle simulations. Code is available at:\\
https://github.com/pranavc2255/gaussian-splatting-novel-view-render.git
\end{abstract}


\section{\textbf{Introduction}}
The field of 3D model reconstruction has seen significant advancements in recent years, driven by the need for accurate and efficient techniques in various applications such as extended reality (XR), photogrammetry, Simultaneous Localization And Mapping (SLAM), and autonomous vehicle simulations. Among the prominent methods for 3D reconstruction, Photogrammetry and Gaussian Splatting have emerged as powerful techniques, each with its unique strengths and limitations. Photogrammetry is a technique that uses overlapping photographs taken from different angles to create detailed and accurate 3D models of objects or scenes [1]. Gaussian Splatting models a scene using a collection of Gaussian functions. Each Gaussian is characterized by parameters such as its 3D position, covariance matrix, opacity, color, and spherical harmonics of the color. These parameters are learned from multiple-view images to represent and render the scene accurately [2].

Novel view synthesis (NVS) refers to generating new, unseen views of a scene that are different from views that were captured to generate the 3D model [3]. This process is crucial for applications requiring dynamic and interactive visual experiences. NeRF has revolutionized the field of novel view synthesis by utilizing neural network to represent a volumetric scene. This method optimizes a continuous volumetric scene function, which can generate high-quality images from novel viewpoints.

In addition to NeRF-based methods, other innovative approaches have been developed to advance NVS. For example, the work by Flynn et al. on DeepView synthesizes novel views by learning a multi-plane image (MPI) representation. This method captures the scene structure and appearance in a layered format, allowing for efficient rendering of novel views by compositing these layers from different viewpoints [4].

Another significant contribution is the Multiview Neural Surface Reconstruction by Srinivasan et al., which improves the geometric accuracy of synthesized views. This method leverages multi-view consistency and neural rendering techniques to produce high-fidelity novel views, addressing some of the limitations of previous NVS approaches[5].

Dynamic view synthesis has also seen notable advancements. For instance, Li et al. introduced Neural 3D Video Synthesis, which extends NVS to dynamic scenes. This approach employs a neural network to learn the temporal dynamics of a scene, enabling the synthesis of novel views over time and capturing motion and changes within the scene [6].

Recent research has also focused on improving the realism and quality of synthesized views. Zhang et al. proposed a method called Neural Sparse Voxel Fields (NSVF), which combines the benefits of voxel-based representations with neural rendering. This approach enhances the rendering quality and efficiency, particularly for scenes with complex geometry and textures [7].

Additionally, the approach by Niemeyer et al., known as GIRAFFE, enables controllable NVS by learning a generative model that allows for explicit control over the scene layout and object appearances. This method facilitates the creation of novel views with specific attributes, providing greater flexibility and control in the synthesis process [8].

Despite the advancements in this field, there are few studies comparing the ability of photogrammetry and Gaussian Splatting to generate high-quality novel views or their performance when using augmented datasets This report presents a detailed comparative analysis of Photogrammetry and Gaussian Splatting. The study begins with creating 2 datasets (indoor and outdoor) of images from a real-world scene, followed by constructing 3D models using both methods. Performance evaluation is conducted using metrics such as structural similarity index (SSIM), peak signal-to-noise ratio (PSNR), learned perceptual image patch similarity (LPIPS), and lp/mm resolution with the help of USAF resolution chart.

A significant contribution of this work is the creation of an enhanced Gaussian Splatting repository. This enhancement enables the rendering of images from novel camera poses generated in the Blender environment, showcasing the potential of Gaussian Splatting in synthesizing high-quality novel views. Additionally, an augmented dataset comprising original ground images and novel views synthesized via Gaussian Splatting is utilized to generate a new photogrammetry model. This augmented photogrammetry model is then compared with the original model, highlighting the benefits of incorporating novel views synthesized from Gaussian Splatting.

The findings of this report provide valuable insights into the comparative strengths and limitations of Photogrammetry and Gaussian Splatting, offering guidance for their application in various domains. The results demonstrate the efficacy of Gaussian Splatting in generating high-quality novel views and its potential for enhancing photogrammetry based 3D reconstructions. This work contributes to the ongoing development of 3D reconstruction techniques and their application in emerging technologies.

\section{\textbf{Methodology}}

\subsection{Dataset Collection}

Two datasets were used for this project: an indoor dataset consisting of images of objects at home, and an outdoor dataset featuring images of rocks at the Walton Centre on the ASU Tempe Campus. The outdoor dataset, provided by my professor, was captured using a Canon EOS R camera, producing images with a 6240 x 4160 pixels resolution. The indoor dataset was captured with an iPhone 13 in HEIC format, which was later converted to JPG format with a resolution of 1600 x 1200 pixels using a HEIC to JPG converter. With these prepared datasets, we used COLMAP, an open-source Structure-from-Motion (SfM) pipeline, to create a sparse point cloud and a Poisson mesh. The sparse point cloud generated from COLMAP was then used for Gaussian Splatting and the poison mesh was imported into Blender to generate camera poses for novel view synthesis. These high-quality 3D models served as the foundation for the subsequent stages of the project, including model training and novel view synthesis.

\begin{figure}[H]
    \centering
    \includegraphics[width=1\linewidth]{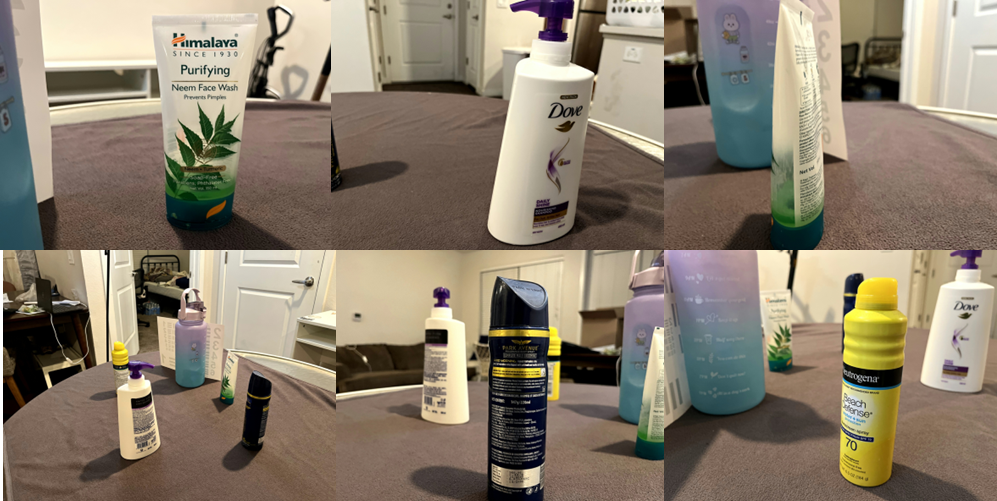}
    \caption{Few images from the Indoor Dataset}
    \label{fig:1}
\end{figure}
\begin{figure}[h]
    \centering
    \includegraphics[width=1\linewidth]{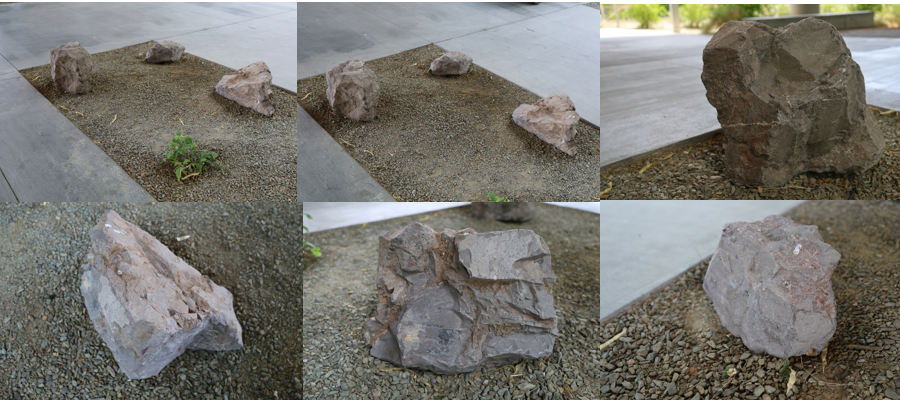}
    \caption{Few images from the Outdoor Dataset}
    \label{fig:enter-label}
\end{figure}

\subsection{3D model construction}
In this phase, I constructed 3D models using both Photogrammetry and 3D Gaussian Splatting based on the ground truth camera poses obtained from the dataset. For photogrammetry, I utilized Agisoft Metashape. The process began with importing the sequentially numbered JPG images into Metashape. The photos were then aligned through feature detection and matching, generating a sparse point cloud and calculating the camera poses for each image. After ensuring the sparse point cloud was accurate, a depth map was created by analyzing each photo more closely. This depth map was then converted into a mesh, which was subsequently textured using the original images to create a photorealistic 3D model [9].

For 3D Gaussian Splatting, I  used the pipeline available on the GitHub
repository [10]. The process begins with using the COLMAP library to create a point cloud from images, followed by generating a dense point cloud to create a Poisson mesh 3D model. Each point in the sparse point cloud is then converted into a Gaussian splat, characterized by attributes such as position, orientation, scale, opacity, and color. The Gaussian splats undergo training with a neural network using Stochastic Gradient Descent to optimize their parameters, minimizing the loss between the rasterized and actual images. Finally, these splats are projected into 2D space and organized by depth through differentiable Gaussian rasterization, enabling accurate and efficient rendering [11].

\begin{figure}[h]
    \centering
    \includegraphics[width=1\linewidth]{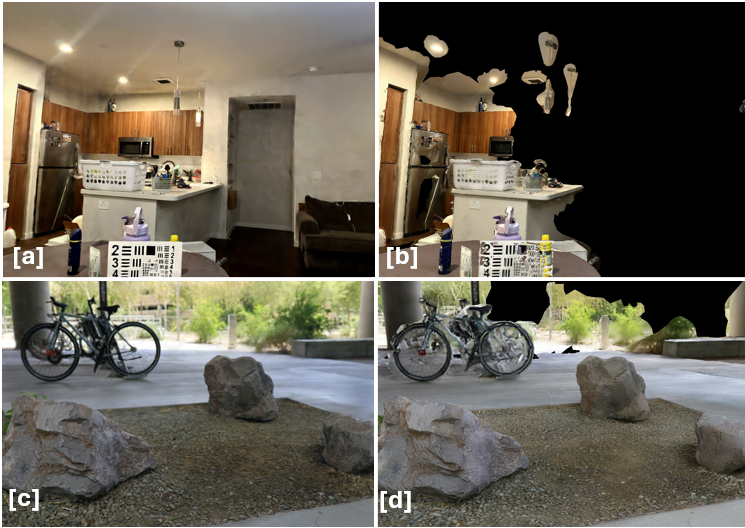}
    \caption{3D Reconstructed models- a) Gaussian Splatting for indoor dataset b) Photogrammetry for indoor dataset c) Gaussian splatting for outdoor dataset d) Photogrammetry for outdoor dataset}
    \label{fig:enter-label}
\end{figure}
The training process was conducted on Google Colab, leveraging its GPU capabilities to accelerate computation. The environment setup included installing essential libraries such as PyTorch, NumPy, and OpenCV, with specific versions chosen to ensure compatibility with the code base. The hardware utilized was Google Colab’s T4 GPU, configured to maximize performance during training. Additionally, several key tools and libraries were employed: Python served as the primary programming language for data processing and analysis, PyTorch was utilized for implementing and training the neural network models, and CUDA was leveraged for custom GPU kernel operations to accelerate rendering processes. Other tools included COLMAP for generating sparse point clouds through Structure-from-Motion (SfM), Agisoft Metashape for photogrammetric modeling and dense cloud generation, OpenCV for various image processing tasks.

\subsection{Novel view synthesis}
The original GitHub repository for Gaussian Splatting restricts rendering images to the initial camera poses used to create the dataset. My current work aims to overcome this limitation by enabling rendering from any defined camera poses within the environment. This is achieved by modifying the repository, which initially relies on camera angles and information generated during the Structure from Motion (SfM) process using COLMAP and is stored in the `images.bin` and `cameras.bin` files.

The crucial camera information for rendering includes quaternion components, translation vectors, image width and height resolutions, focal length, and camera center, primarily defined in \texttt{camera.py} within the \texttt{scenes\_nv} folder. Focal lengths are retrieved using the \texttt{readColmapCameras} function in \texttt{dataset\_readers.py}. To enable novel view synthesis, I modified the repository files to unlink the dependencies on \texttt{cameras.bin} and \texttt{images.bin}, allowing user-defined camera poses for rendering.

\begin{figure}[H]
    \centering
    \includegraphics[width=1\linewidth]{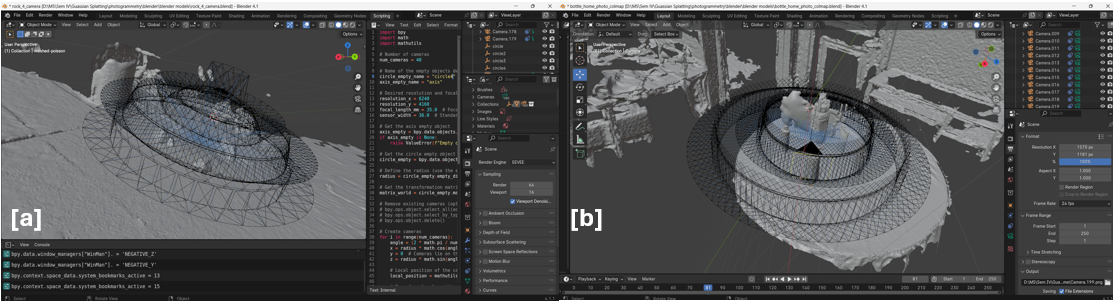}
    \caption{Camera poses created in Blender on circular paths}
    \label{fig:enter-label}
\end{figure}

I created a new process to define customized camera poses using Blender. This involved developing a Python script that adds cameras with the required resolutions and focal lengths in a defined radius centered around the imported 3D model from photogrammetry. The script allows customization of the number of cameras and their positions, enabling multiple rings of cameras at different elevations or positions for varied perspectives. When importing the COLMAP .ply mesh into Blender, it is essential to select the -ve Z forward and -ve Y upward coordinate system directions. Once the camera angles are set, the file is exported from Blender in the .fbx format, ensuring the -ve Z forward and +ve Y upward directions, which is the default setting for exporting but needs verification.

The camera poses from the .fbx file are then converted to COLMAP poses using the camorph library [12], due to differences in coordinate systems. This conversion process results in two files: \texttt{cameras.txt} and \texttt{images.txt}. The \texttt{images.txt} file initially contains 2D point information of the ground truth, which is irrelevant to novel camera poses. To address this, we convert the \texttt{images.txt} into a new modified version using the \texttt{convert\_images\_txt.py} script. The new \texttt{images.txt}, along with \texttt{cameras.txt} and \texttt{point\_cloud.py}, serve as inputs for our novel view synthesis, as depicted in the flowchart.
\begin{figure*}[h]
\centering
\includegraphics[width=\textwidth]{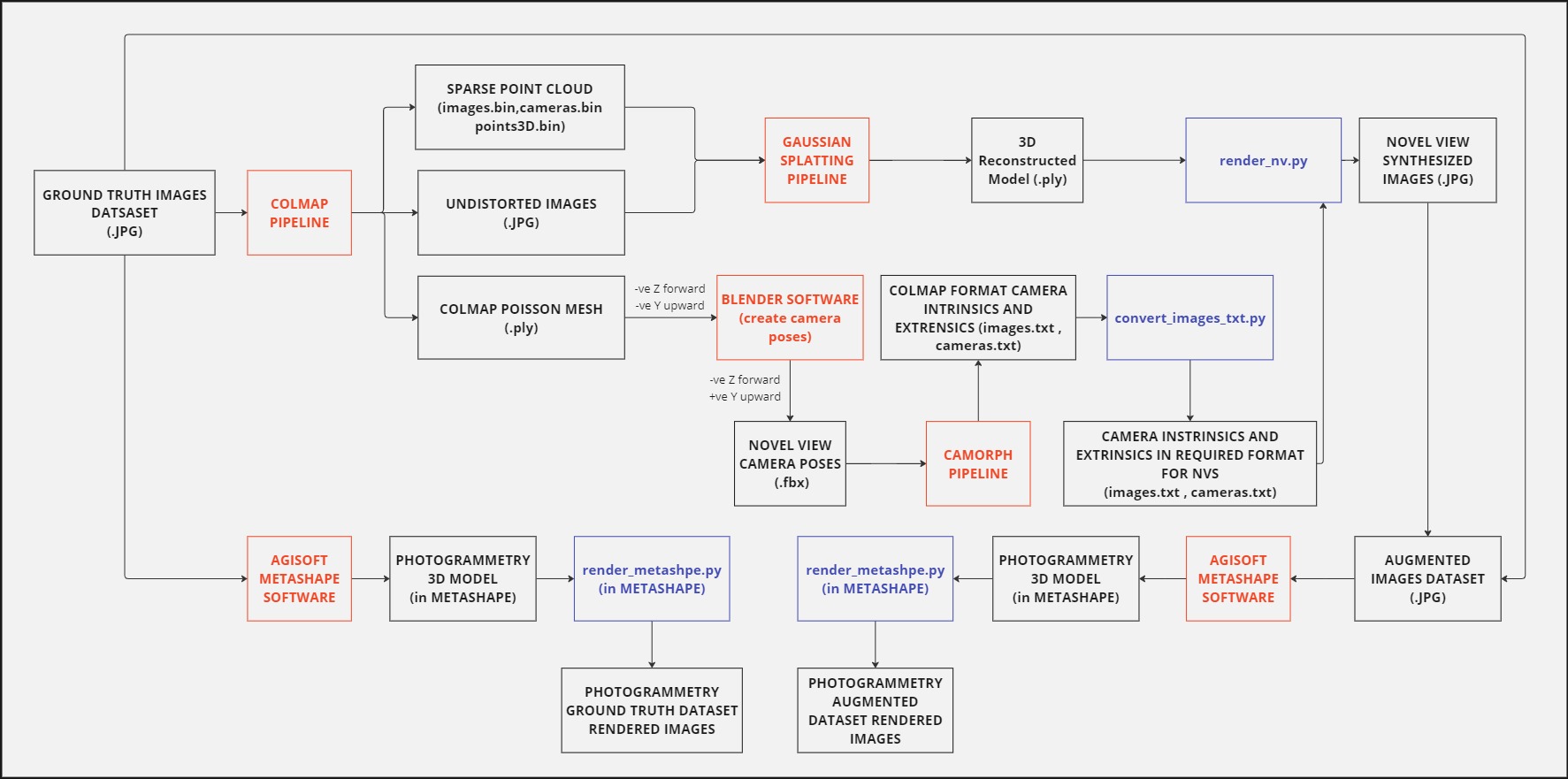} 
\caption{Flowchart depicting the process of novel view synthesis and Photogrammetry with Gaussian Splatting NVS augmented dataset}
\label{fig:flowchart}
\end{figure*}
 
This approach ensures that we can render images from any defined camera pose, significantly enhancing the flexibility and applicability of the Gaussian Splatting method. The flowchart outlines the detailed process, including the generation of novel view synthesized images and the creation of an augmented dataset comprising original ground truth images and novel views for photogrammetry.

The new \texttt{images.txt}, along with \texttt{cameras.txt} and \texttt{point\_cloud.py}, serve as inputs for our novel view synthesis. This approach ensures that we can render images from any defined camera pose, significantly enhancing the flexibility and applicability of the Gaussian Splatting method.
\vspace{-22pt}
\subsection{Augmented Dataset Creation}
An augmented dataset was created by combining the original ground images with novel views synthesized using the Gaussian Splatting method. This dataset was then used to generate an augmented dataset photogrammetry model.

\begin{figure} [H]
    \centering
    \includegraphics[width=1\linewidth]{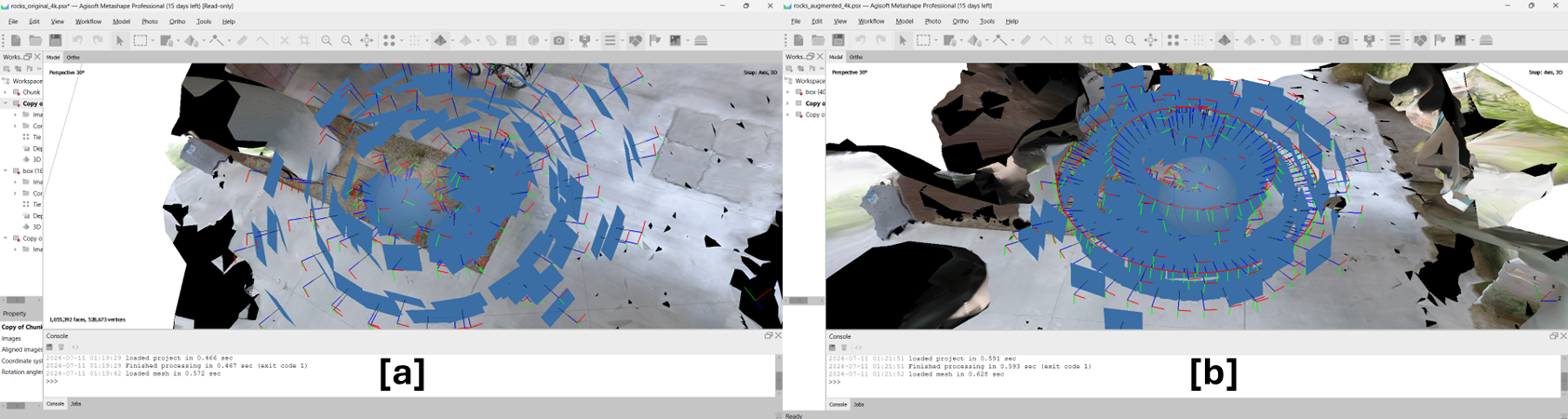}
    \caption{Metashape Camera poses for a) Ground Truth dataset b) Augmented dataset }
    \label{fig:enter-label}
\end{figure}

\subsection{Comparison}

As part of this work, three key comparisons were conducted to evaluate the effectiveness of the augmented dataset and different modeling techniques:
\begin{enumerate}
    \item Comparison between Gaussian Splatting and Photogrammetry Models Created from Original Camera Poses for both datasets (Parameters- SSIM, LPIPS, PSNR)
    \item Comparison between Photogrammetry Models Created from Original Camera Poses and Augmented Camera Poses for both datasets (Parameters- SSIM, LPIPS, PSNR)
     \item Comparison of spatial resolutions of Gaussian Splatting and Photogrammetry Models Created from Original Camera Poses for indoor dataset (Parameters- 1951 USAF Resolution Chart)
    \item Comparison of spatial resolutions of Photogrammetry Models Created from Original Camera Poses and Augmented for indoor dataset (Parameters- 1951 USAF Resolution Chart)
    
\end{enumerate}
Parameters of Comparison
\begin{enumerate}
    \item Structural Similarity Index (SSIM): Measures the structural similarity between ground truth images and rendered images, with higher values indicating better preservation of image details and structures. The formula used for calculating SSIM is -
\begin{equation}
\text{SSIM}(x, y) = \frac{(2 \mu_x \mu_y + C_1)(2 \sigma_{xy} + C_2)}{(\mu_x^2 + \mu_y^2 + C_1)(\sigma_x^2 + \sigma_y^2 + C_2)}
\end{equation}

Where:
\begin{align*}
\mu_x & \text{ and } \mu_y \text{ are the average of } x \text{ and } y. \\
\sigma_x^2 & \text{ and } \sigma_y^2 \text{ are the variance of } x \text{ and } y. \\
\sigma_{xy} & \text{ is the covariance of } x \text{ and } y. \\
C_1 &= (K_1 L)^2 \\
C_2 &= (K_2 L)^2 \\
L & \text{ is the dynamic range of the pixel-values}  \\
K_1 &= 0.01 \\
K_2 &= 0.03
\end{align*}
    \item Peak Signal-to-Noise Ratio (PSNR): Quantifies the reconstruction quality of the images, measuring the ratio between the maximum possible power of a signal and the power of corrupting noise, with higher values indicating better image quality. The formula to calculate PSNR is -
\begin{equation}
\text{PSNR} = 10 \log_{10} \left( \frac{R^2}{\text{MSE}} \right)
\end{equation}
Where:
\begin{equation}
\text{MSE} = \frac{1}{M \cdot N} \sum_{m=0}^{M-1} \sum_{n=0}^{N-1} [I_1(m,n) - I_2(m,n)]^2
\end{equation} \\
\begin{itemize}
   
    \item \( I_1(m,n) \) and \( I_2(m,n) \) are the pixel values of the original and compressed images, respectively.
    \item \( M \) and \( N \) are the dimensions of the images.
    \item \( R \) is the maximum possible pixel value of the images.
\end{itemize}

    \item Learned Perceptual Image Patch Similarity (LPIPS): Evaluates the perceptual similarity between ground truth and rendered images using deep learning features, with lower values indicating better perceptual quality. LPIPS is calculated by passing the images through a pre-trained deep neural network, extracting feature representations from intermediate layers, and then computing the distance between these features using a learned linear weighting. 
    \item Spatial Resolution(lp/mm using USAF resolution chart): Assesses the line pairs per millimeter (lp/mm) that can be resolved in the images, providing an objective measure of the models' resolution capabilities. The formula for calculating the USAF resolution is - 
\begin{equation}
\text{Resolution} = 2^{\left(\text{Group Number} + \frac{\text{Element Number} - 1}{6}\right)}
\end{equation}

Where:
\begin{itemize}
    \item Group Number: The group in the USAF resolution test chart, which determines the scale of the test pattern.
    \item Element Number: The specific element within the group, each group contains 6 elements.
\end{itemize}
\end{enumerate}

\section{\textbf{Results}}
As part of this work, four key comparisons were conducted to evaluate the effectiveness of the augmented dataset and different modeling techniques.
\subsection{ Comparison between Gaussian Splatting and Photogrammetry Models Created from Original Camera Poses for both datasets}
 
\begin{figure}[H]
    \centering
    \includegraphics[width=1\linewidth]{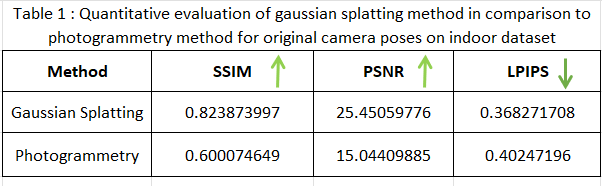}
    \label{fig:enter-label}
\end{figure}

These results indicate that Gaussian Splatting achieves a significantly higher score in every metric - SSIM, PSNR, and LPIPS  for the indoor dataset (See Table 1). The considerable improvement in SSIM and PSNR suggests that Gaussian Splatting captures more detailed and accurate reconstructions, while the lower LPIPS value indicates better perceptual similarity to the ground truth images.
\begin{figure}[H]
    \centering
    \includegraphics[width=1\linewidth]{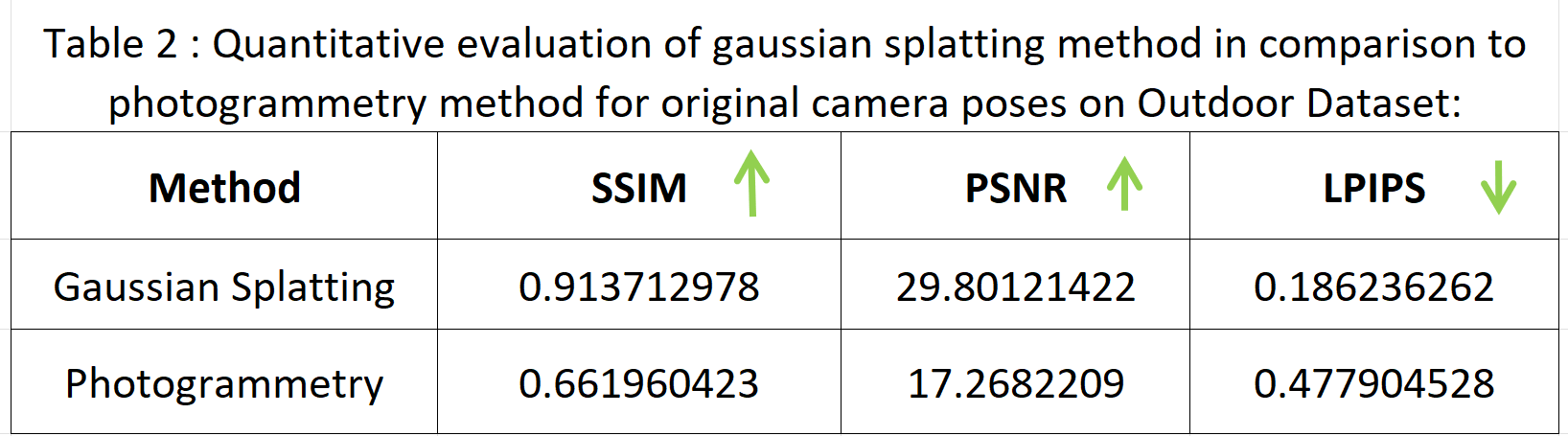}
    \label{fig:enter-label}
\end{figure}
\vspace{-20pt}
Similarly, for the outdoor dataset, Gaussian Splatting again demonstrated superior performance (See Table 2). The results for the outdoor dataset show an even larger performance gap between Gaussian Splatting and Photogrammetry. The higher SSIM and PSNR values indicate that Gaussian Splatting provides more accurate and detailed reconstructions, while the significantly lower LPIPS value shows much better perceptual quality.

\subsection{ Comparison between Photogrammetry Models Created from Original Camera Poses and Augmented Camera Poses for both datasets}
\begin{figure}[H]
    \centering
    \includegraphics[width=1\linewidth]{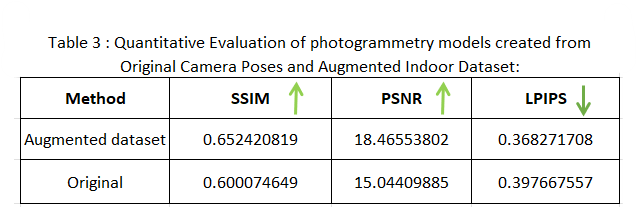}
    \label{fig:enter-label}
\end{figure}
\vspace{-10pt}
For the indoor dataset, the augmented dataset model achieved a higher SSIM compared to the original model, indicating better preservation of image details and structures. The PSNR value for the augmented dataset was significantly higher compared to the original dataset, signifying improved image quality. Additionally, the LPIPS value for the augmented dataset was lower compared to the original dataset, demonstrating better perceptual quality. (See Table 3)

Similarly, for the outdoor dataset, the augmented dataset model showed a higher SSIM compared to the original model. The PSNR value for the augmented dataset was also higher compared to the original dataset. The LPIPS value for the augmented dataset was slightly higher compared to the original dataset, indicating a marginal decrease in perceptual quality.(See Table 4).
\begin{figure}[H]
    \centering
    \includegraphics[width=1\linewidth]{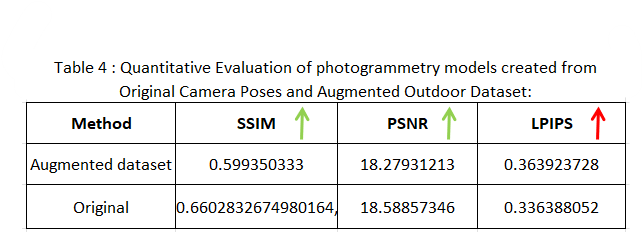}
    \label{fig:enter-label}
\end{figure}

\subsection{Comparison of spatial resolutions of Gaussian Splatting and Photogrammetry Models Created from Original Camera Poses for indoor dataset}
In this study, we evaluated the resolution of 3D models created using Gaussian Splatting and photogrammetry techniques on an indoor dataset, which included a 1951 USAF resolution chart for precise measurement. Two models were constructed: one using Gaussian Splatting and the other using photogrammetry.

\begin{figure} [H]
    \centering
    \includegraphics[width=1\linewidth]{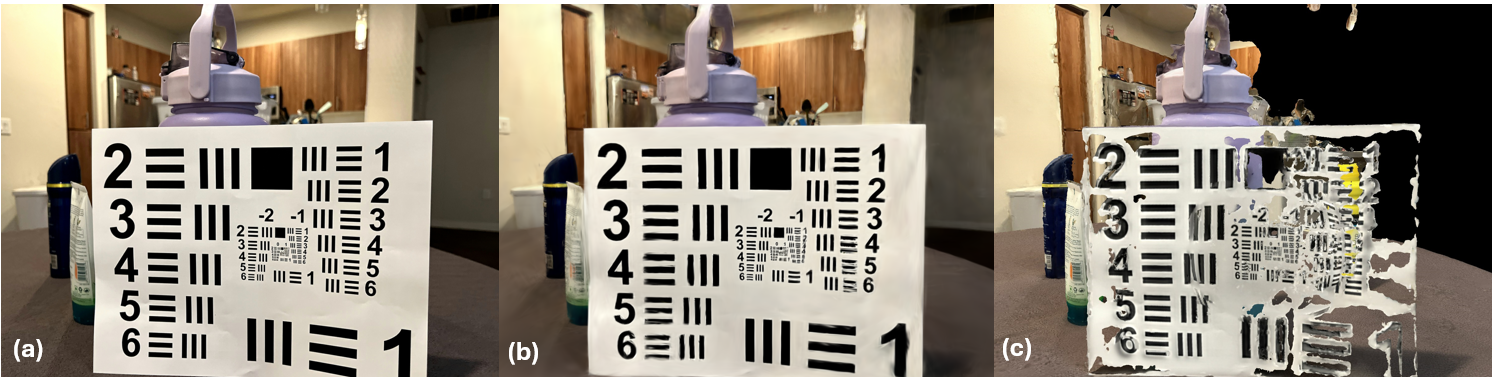}
    \caption{Analysis of spatial resolution using 1951 USAF resolution chart a) Ground truth b) Gaussian Splatting (original indoor dataset) c) Photogrammetry (original indoor dataset) }
    \label{fig:enter-label}
\end{figure}

The Gaussian Splatting model achieved a resolution of 0.707 lp/mm, indicating a higher level of detail and sharpness in the reconstructed model. In comparison, the photogrammetry model demonstrated a resolution of 0.561 lp/mm (See Table 5 and Fig. 7). This result highlights a significant difference in the level of detail captured by the two methods, with Gaussian Splatting outperforming photogrammetry in terms of resolution.

\subsection{Comparison of spatial resolutions of Photogrammetry Models Created from Original Camera Poses and Augmented for indoor dataset}

In this study, I investigated the impact of augmenting a photogrammetry dataset with Gaussian Splatting rendered images on 3D model resolution and completeness. Using 225 ground truth images, we created a photogrammetry model in Metashape. Combining these with 100 GSplat-rendered images, we formed an augmented dataset of 325 images to create a new photogrammetry model (See Table 5 and Fig. 8).
\begin{figure}[H]
    \centering
    \includegraphics[width=1\linewidth]{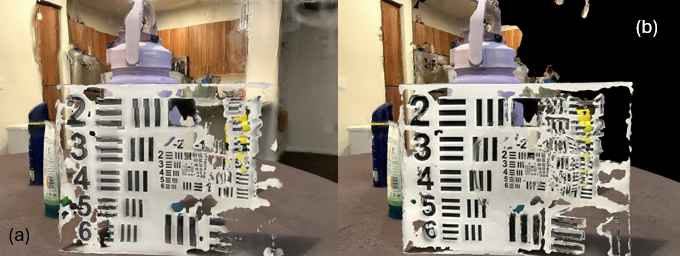}
    \caption{Analysis of spatial resolution using 1951  USAF resolution chart on rendered images using Photogrammetry of a) augmented dataset b) original dataset}
    \label{fig:enter-label}
\end{figure}
\begin{figure}
    \centering
    \includegraphics[width=1\linewidth]{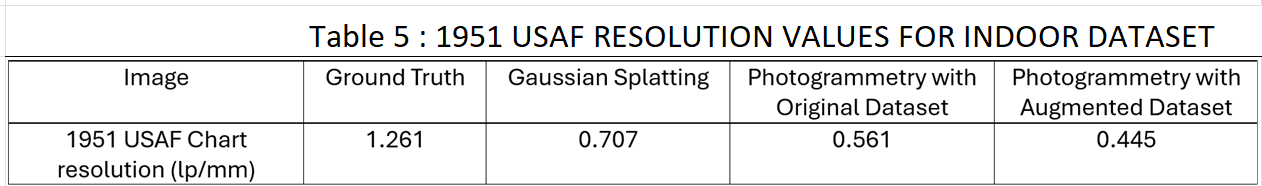}
    \label{fig:enter-label}
\end{figure}
The original photogrammetry model, created using 225 ground truth images, exhibited a resolution of 0.561 lp/mm as measured by the USAF 1951 resolution chart. In contrast, the augmented photogrammetry model, constructed from 325 images, showed a lower resolution of 0.445 lp/mm. The GSplat-rendered images, particularly those capturing the 1951 USAF resolution chart from non-perpendicular angles, were found to be blurry and noisy. This noise adversely affected the resolution of the photogrammetry model. Despite the reduction in resolution, the augmented dataset resulted in a more complete reconstruction of the scene. The additional GSplat images provided supplementary information about the background and occluded areas, enhancing the overall understanding of the scene.

\begin{figure} [H]
    \centering
    \includegraphics[width=1\linewidth]{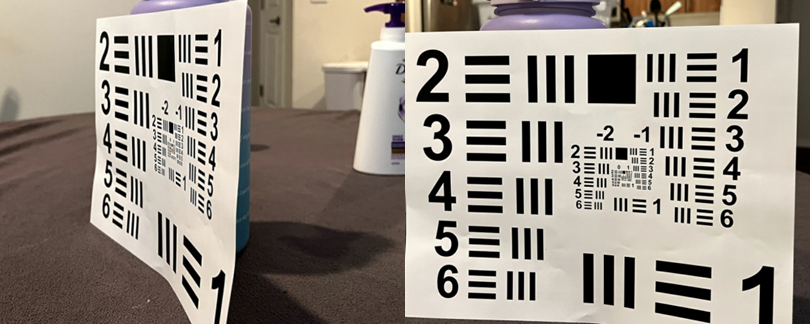}
    \caption{Ground truth images with no noise}
    \label{fig:enter-label}
\end{figure}
\vspace{-15pt}
\begin{figure}[H]
    \centering
    \includegraphics[width=1\linewidth]{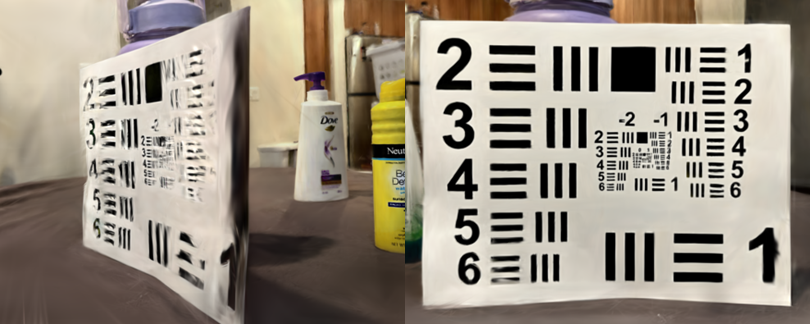}
    \caption{Noise introduced in Gaussian Splatting NVS image when rendered from pose very different from ground truth camera pose}
    \label{fig:enter-label}
\end{figure}

\section{\textbf{Conclusion}}

The comparative analysis of photogrammetry models created with just ground truth images and photogrammetry models created with Gaussian Splatting (GSplat) novel view synthesized augmented images reveals significant insights into the advantages and challenges of using augmented datasets. The findings demonstrate that photogrammetry models created from augmented datasets consistently outperform those created from original datasets in terms of SSIM and PSNR across both indoor and outdoor scenarios. This superior performance indicates that augmented datasets enhance the preservation of image details and overall image quality.

However, the perceptual quality, measured by the LPIPS, presents mixed results. For the indoor dataset, the augmented dataset model exhibited a lower LPIPS value, indicating better perceptual quality. Conversely, for the outdoor dataset, the LPIPS value was slightly higher for the augmented dataset model, suggesting a minor decrease in perceptual quality. These variations highlight the need for further refinement in the augmentation process to consistently enhance perceptual quality across different environments.

The integration of GSplat-rendered images into the photogrammetry workflow offers notable benefits, such as improved scene coverage by providing additional details about occluded areas and the background. However, it also introduces challenges, particularly in the form of noise that can reduce the model's resolution. Our study found that novel view synthesized images created from camera poses closer to the original camera poses resulted in negligible noise, while those created from poses further away introduced significant noise into the rendered images. This underscores the importance of optimizing the placement of novel camera poses.

Instead of creating cameras on a circular path in Blender, future work should focus on developing a method to create camera poses at some offset distance on all sides from the original camera poses to determine what optimizes the quality. Ensuring that novel camera poses are not excessively distant from the original ones will help maintain the quality of NVS images, thereby enhancing both the resolution and completeness of the reconstructed 3D models. Advanced augmentation techniques and optimization of the Gaussian splatting process are potential avenues for further improving the perceptual quality and overall effectiveness of augmented datasets in photogrammetry.
\section{\textbf{Acknowledgements}}
 \setlength{\parindent}{1.6cm} I would like to express my deepest gratitude to \textbf{Dr. Jnaneshwar Das, Assistant Research Professor at the School of Earth and Space Exploration, Arizona State University, Tempe,} for his invaluable guidance and support throughout this applied project. His expertise and insights were instrumental in the successful completion of this work. I would also like to thank him for providing the outdoor dataset featuring images of rocks at the Walton Centre on the ASU Tempe Campus used in this study.

\end{document}